\documentclass[journal]{IEEEtran}

\ifCLASSINFOpdf
\else
   \usepackage[dvips]{graphicx}
\fi
\usepackage{url}

\usepackage{graphicx}
\usepackage{amsmath,epsfig}
\usepackage{amssymb}
\usepackage{booktabs}
\usepackage{float}
\usepackage{bm}
\usepackage{booktabs}
\usepackage{multirow}
\usepackage{makecell}
\usepackage{color}
\usepackage{algorithm}
\usepackage{algorithmicx}

\usepackage{algpseudocode}
\usepackage{subfigure}

\begin{document}

\title{Shaping Visual Representations with Attributes for Few-Shot Recognition}

\author{Haoxing~Chen,~\IEEEmembership{}
	Huaxiong~Li,~\IEEEmembership{}
	Yaohui~Li,~\IEEEmembership{}
	Chunlin~Chen~\IEEEmembership{}
\thanks{This work was partially supported by the National Natural Science Foundation of China (Nos. 62176116, 62073160, 71732003), and the Natural Science Foundation of the Jiangsu Higher Education Institutions of China, No. 20KJA520006.
	
H. Chen, H. Li, Y. Li and C. Chen are with the Department of Control and Systems Engineering, Nanjing University, Nanjing 210093, China (e-mail: haoxingchen@smail.nju.edu.cn; huaxiongli@nju.edu.cn; yaohuili@smail.nju.edu.cn; clchen@nju.edu.cn).}}

\maketitle

\begin{abstract}
Few-shot recognition aims to recognize novel categories under low-data regimes. Some recent few-shot recognition methods introduce auxiliary semantic modality, i.e., category attribute information, into representation learning, which enhances the feature discrimination and improves the recognition performance. Most of these existing methods only consider the attribute information of support set while ignoring the query set, resulting in a potential loss of performance. In this letter, we propose a novel attribute-shaped learning (ASL) framework, which can jointly perform query attributes generation and discriminative visual representation learning for few-shot recognition. Specifically, a visual-attribute predictor (VAP) is constructed to predict the attributes of queries. By leveraging the attributes information, an attribute-visual attention module (AVAM) is designed, which can adaptively utilize attributes and visual representations to learn more discriminative features. Under the guidance of attribute modality, our method can learn enhanced semantic-aware representation for classification. Experiments demonstrate that our method can achieve competitive results on CUB and SUN benchmarks. Our source code is available at: \url{https://github.com/chenhaoxing/ASL}.
\end{abstract}


\IEEEpeerreviewmaketitle

\section{Introduction}
Deep learning has achieved outstanding performance in many visual tasks~\cite{li2021adaptive,Gao_2021_ICCV,HashFormer,zl_ag}. However, training deep models often requires a lot of labeled data, which is not always accessible in real applications~\cite{select_eccv20,LoFGAN,ifsm,das2022confess}. Inspired by the ability of humans, few-shot learning aims to recognize new objects with few labeled training samples.

Recently, various successful few-shot recognition methods have been proposed, which can be roughly divided into three categories: meta-learning methods~\cite{finn2017model,ren18iclr}, data augmentation methods~\cite{AGA} and metric-learning based methods~\cite{snell2017prototypical,chen2021ssformers,bmpn}. In these methods, metric-learning based methods have attracted extensive attention due to their simplicity and effectiveness. Most of them mainly focus on enhancing the informativeness and discriminability of learned semantic visual representations to improve image recognition accuracy. However, these methods perform image classification in the context of uni-modal visual learning, i.e., only using the images. In the real-world, humans learn new concepts by leveraging multi-modal information rather than a single one~\cite{clip,gzrl}. For instance, one can learn about the \textit{red billed blue magpie} by some images and attributes, such as \textit{red bill} and \textit{red nape}, instead of only seeing many pictures. \textit{In other words, attribute information can help humans learn new visual objects}~\cite{comp,transzero,mm2020Learning,zsl_rlr}.

\begin{figure}[t]
	\centering
	\includegraphics[height=2.5cm,width=8.5cm]{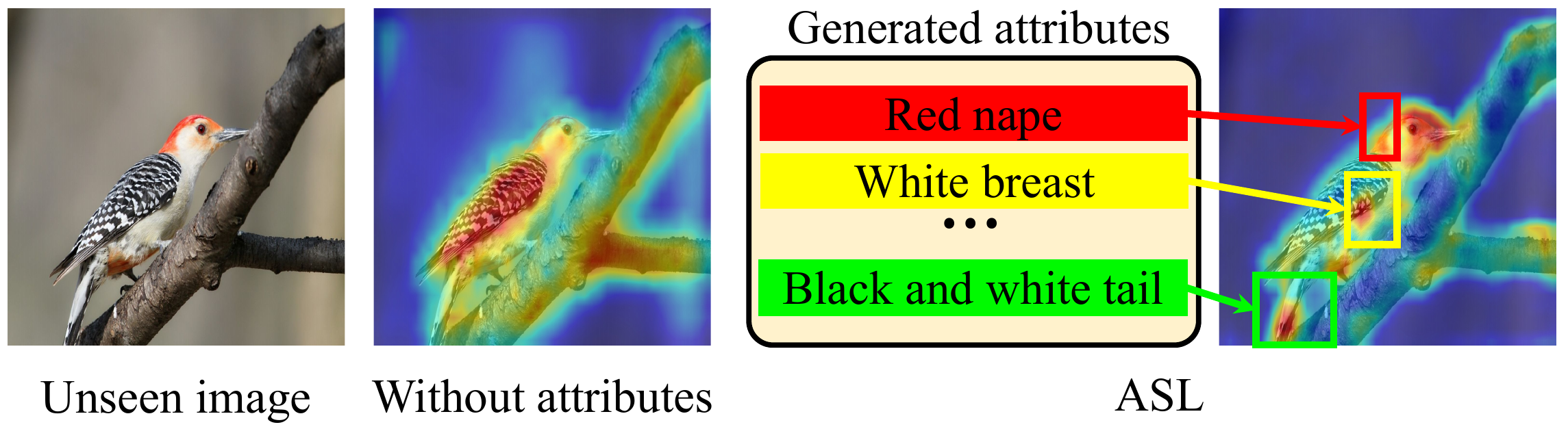}
	\caption{An illustration of the effect of our proposed ASL method. }
	\label{fig1}
\end{figure}

To imitate this ability, some works introduce auxiliary semantic modalities to enhance feature learning in few-shot learning. Pavel \textit{et al.}~\cite{comp} took attribute features as supplementary information and enhanced the representation learning ability by adding regularization terms. AM3~\cite{acm} utilized convex combination to adaptively mix the semantic structures of visual representations and label semantics. Dual TriNet~\cite{DualTriNet} is an auto-encoder network, which directly synthesizes instance features by leveraging semantics.

Although the above methods~\cite{comp,acm,DualTriNet} have achieved impressive performance by leveraging semantic information, they only learn the feature representations of support set with the help of attribute information and ignore the query set that lacks the attribute information. To the best of our knowledge, there is no specific mechanism to explore the underlying attribute information of query samples, which is hopeful to improve the few-shot recognition performance. 

Towards this end, we propose a novel attribute-shaped learning (ASL) model that generates corresponding attributes through image features and learns more discriminative visual features combined with attributes. Specifically, since query images have no auxiliary attribute information, a visual-attribute predictor (VAP) is proposed to generate attribute features. Then, we propose an attribute-visual attention module (AVAM), which can adaptively utilize attributes and visual representations to learn more discriminative features. AVAM contains two sub-modules, i.e., channel attention module (CAM) and pyramid spatial attention module (PSAM). CAM finds important channels, and PSAM finds multi-scale spatial-visual representations. Fig. 1 shows the effectiveness of our ASL method and classical ProtoNets~\cite{snell2017prototypical}. It can be observed that, by incorporating the generated attribute information, our method can focus on more discriminative local regions for visual representation learning, contributing to better recognition performance. To summarize, our main contributions are as follows: 1) We propose a novel visual-attribute predictor (VAP), which can generate attributes for query images to assist the learning of visual representations. 2) We propose an attribute-visual attention module (AVAM), which can simultaneously leverage attribute and visual information to adaptively learn the discriminative features in images. 3) We show that our method achieves competitive results compared to other state-of-the-art methods.

\section{METHODOLOGY}
\begin{figure}[t]
	\centering
	\includegraphics[height=9.3cm,width=8cm]{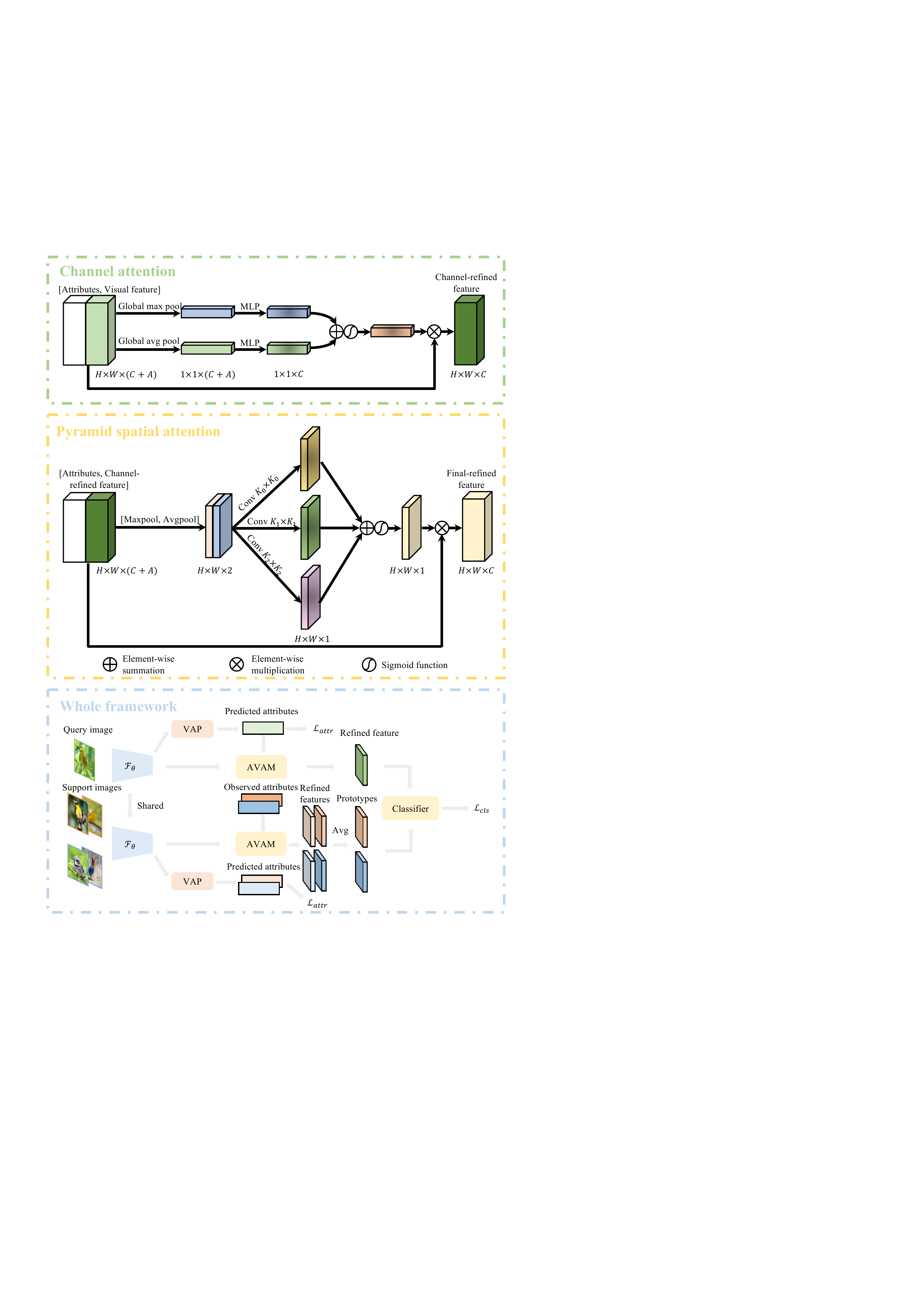}
	\caption{Illustration of attribute-visual attention module }
	\label{arc}
\end{figure}

In few-shot learning, the model is trained by a series of $N$-way $M$-shot episodes and each episode can be seen as an independent task. Each task $\tau$ is formed by randomly selecting $N$ categories from training set $\mathcal{D}_{train}$, and then sampling support set $\mathcal{S}=\{(s_i, a_i, y_i)\}^{N\times M}_{i=1}$ and query set $\mathcal{Q}=\{(q_j, a_j, y_j)\}^{Q}_{j=1}$ from these $N$ categories. Here, $x$, $a\in\mathbb{R}^A$ and $y$ represent the image, attribute vector and label respectively. Note that $\mathcal{Q}$ contains different examples from the same $N$ categories and $A$ is determined by the dataset.
After learning on trainning set $\mathcal{D}_{train}$, a model is evaluated on a test set $\mathcal{D}_{test}$.

The proposed method is applicable to nearly any metric-learning based framework. Since ProtoNets~\cite{snell2017prototypical} is simple and effective, we choose it as our baseline framework. ProtoNets aim to generate the prototype for each category and determine the category of query samples by calculating the distances to each prototype. Specifically, ProtoNets firstly obtain the feature representations of each image through a feature extractor $\mathcal{F}_{\theta}$, and then compute prototype a $p_n$ for support class $n$:
\begin{equation}
	p^n = \frac{1}{M} \sum_{i=1}^{M} \mathcal{F}_{\theta}(s^n_i),
\end{equation}
where $s^n_i$ is the $i$-th image of support category $n$.
For each query image $(q_j,y_j)$, ProtoNets calculate the Euclidean distance between $q$ and each prototype, and use softmax function to obtain the probability distribution:
\begin{equation}
	p(\hat{y}_j=n|q_j) = \frac{{\rm exp}(-||\mathcal{F}_{\theta}(q_j)-p^n||^2)}{\sum_{k=1}^{N}{\rm exp}(-||\mathcal{F}_{\theta}(q_j), p^k||^2)},
\end{equation}
$\mathcal{F}_{\theta}$ is trained by minimizing the classification loss:
\begin{equation}
	\mathcal{L}_{cls} = -\sum_{j=1}^{Q}\log p(\hat{y}_{j}=y_{j}|q_j).
\end{equation}
\subsection{Visual-Attribute Predictor}
To enable both the query images and support categories to have additional attribute descriptions, we generate attribute descriptions for each sample. Specifically, given an image $x$, through $\mathcal{F}_{\theta}$, we can get visual representation $\mathcal{F}_{\theta}(x)\in\mathbb{R}^{H\times W\times C}$, where $W$, $H$ and $C$ are width, height and channel dimension. We first use global-average pooling (GAP) to get pooled fetures for attribute prediction. Then define a visual-attribute predictor $g_{\phi}$: $\mathbb{R}^C \mapsto \mathbb{R}^{A}$, which can predict $A$ number of attributes. To validate the quality of the generated attribute vector, we define the loss function:
\begin{equation}
	\mathcal{L}_{attr} = - \frac{1}{A} \sum_{i=1}^{N\times K+Q}\sum_{1}^{A}(\hat{a}^k_i-a_i^k),
\end{equation}
where $a^k_i$ is the $i$-th observed attribute of the $k$-th sample and $\hat{a}_i^k$ is the predicted ones. The whole optimization objectives of the whole model are as follows:
\begin{equation}
	\mathcal{L} = 
	\mathcal{L}_{cls} + \alpha \cdot \mathcal{L}_{attr},
\end{equation}
where $\alpha$ is the weighting factor of $\mathcal{L}_{attr}$. The training strategy of ASL is illustrated in Algorithm 1.

\subsection{Attribute-Visual Attention Module}
To use attribute vectors to generate more discriminating features, we propose an attribute-visual attention module (AVAM). As shown in Fig. 2, AVAM consists of two sub-models, i.e., channel attention module (CAM) and pyramid spatial attention module (PSAM). CAM to blend cross-channel information and learn which channels to focus on. And PSAM can extract multi-scale spatial information at a more granular level by attributes guiding. Note that AVAM is a flexible module and can be easily added into any few-shot learning method.

\begin{algorithm}[t]
	\caption{Training strategy of ASL}
	\textbf{Input:} training set $\mathcal{D}_{train}$;
	\begin{algorithmic}[1] 
		\ForAll{iteration=1, ..., MaxIteration}
		\State sample $N$-way $M$-shot task $(\mathcal{Q},\mathcal{S})$ from $\mathcal{D}_{train}$;
		\State compute support visual features $\{F_{\theta}(s_i)_{i=1}\}^{N\times M}$;
		\State compute refined support features Eq. (6)-(9);
		\State compute prototypes by Eq. (1);
		\For{$j$ in $\{1,...,Q\}$}
		\State compute query visual feature $F_{\theta}(q_j)$;
		\State compute refined query features Eq. (6)-(9);
		\State obtain the probability distribution by Eq. (2);
		\EndFor
		\State obtain training loss by Eq. (3)-(5);
		\State update parametres in ASL by Adam;
		\EndFor
	\end{algorithmic}
	\textbf{Output:} trained ASL.
\end{algorithm}

Specifically, we use the attributes provided by the dataset for support images and the generated attributes for query images. Given a feature map $F^v\in\mathbb{R}^{H\times W\times C}$ and attributes vector $F^a\in\mathbb{R}^{A}$, we first broadcast $F^a$ along height and width dimension of $F^v$, and then concatenate $F^v$ and $F^a$ to get hybrid feature $F\in\mathbb{R}^{H\times W\times (C+A)}$. Through CAM, we can get channel-refined feature $F_{c}\in\mathbb{R}^{H\times W\times C}$. Then we broadcast $F^a$ along height and width dimension of $F_c$, and concatenate $F_c$ and $F^a$ to get hybrid feature $F'\in\mathbb{R}^{H\times W\times (C+A)}$. Finally, we feed $F'$ into PSAM to get the final-refined feature $F_{f}\in\mathbb{R}^{H\times W\times C}$.
The overall attribute-visual attention process can be summarized as:
\begin{gather}
	F_{c} = \mathcal{M}_c(F) \otimes F,\\
	F_{f} = \mathcal{M}_s(F')\otimes F_{c},
\end{gather}
where $\otimes$ denotes element-wise multiplication, $ \mathcal{M}_c(\mathcal{X})\in\mathbb{R}^{1\times1\times C}$ denotes channel attention map and $\mathcal{M}_s(\mathcal{X})\in\mathbb{R}^{H \times W\times1}$ denotes spatial attention map.

\textbf{Channel attention.} 
To compute the channel-wise attention efficiently, we first use global average pooling and global max pooling to  aggregate channel information and get $F^{avg}\in\mathbb{R}^{1\times 1\times (C+A)}$ and $F^{max}\in\mathbb{R}^{1\times 1\times (C+A)}$. Then, channel attention generating network (i.e., one layer MLP) is adopted to generate a channel-wise attention map $\mathcal{M}_c$. To summarize, the channel attention is computed as:
\begin{gather}
	\mathcal{M}_c(F) = \sigma({\rm MLP}(F^{avg})+
	{\rm MLP}(F^{max})).
\end{gather}

\textbf{Pyramid spatial attention.} 
Since different visual representation of attributes correspond to different sizes in images~\cite{apcnn}, it is necessary to design a pyramid spatial attention module. Similar to the channel attention module, we first apply average-pooling and max-pooling operations along the channel dimension and concatenate the pooled features. Then we aggregate these two features via $K$ different 2D convolution kernels. Finally, we sum all spatial-wise attention maps to generate the final spatial attention map. The pyramid spatial attention can be described as:
\begin{equation}	
	\mathcal{M}_s(F') = \sigma(\sum_{i=1}^{K}
	{\rm Conv}_{K_i}([{\rm AvgPool}(F');{\rm MaxPool}(F')])),
\end{equation}
where ${\rm Conv}_{K_i}$ represents a 2D convolution operation with the filter size of $K_i\times K_i$.

\begin{table}[t]
	\centering
	\caption{Comparison with other state-of-the-art methods with $95\%$ confidence intervals on CUB and SUN.}
	\begin{tabular}{c c c c }
		\toprule
		\multirow{2}{*}{\textbf{Model}}&\multirow{2}{*}{\textbf{Backbone}}& \multicolumn{2}{c}{\textbf{CUB}}
		\\
		\cline{3-4}
		&& \textbf{1-shot} & \textbf{5-shot}\\
		\midrule
		Matching Nets~\cite{vinyals2016matching}&Conv-64F&  61.16$\pm$\footnotesize{0.89} & 72.86$\pm$\footnotesize{0.70}\\
		ProtoNets~\cite{snell2017prototypical}&Conv-64F&  51.31$\pm$\footnotesize{0.91} & 70.77$\pm$\footnotesize{0.69}\\
		Relation Nets~\cite{sung2018learning}&Conv-64F& 62.45$\pm$\footnotesize{0.98} & 76.11$\pm$\footnotesize{0.69}\\
		Comp.~\cite{comp}  &Conv-64F&53.60$\pm$\footnotesize{0.00} & 74.60$\pm$\footnotesize{0.00}\\
		CovaMNet~\cite{li2019distribution}  &Conv-64F&60.58$\pm$\footnotesize{0.69} & 74.24$\pm$\footnotesize{0.68}\\
		LRPABN~\cite{lrpabn}  &Conv-64F&67.97$\pm$\footnotesize{0.44} & 78.26$\pm$\footnotesize{0.22}\\
		\midrule  
		\textbf{ASL}&Conv-64F& \textbf{74.82$\pm$\footnotesize{0.17}} &\textbf{80.15$\pm$\footnotesize{0.12}}\\
		\midrule
		ProtoNets~\cite{snell2017prototypical}&ResNet-12&  68.80$\pm$\footnotesize{0.00} & 76.40$\pm$\footnotesize{0.00}\\
		Relation Nets~\cite{sung2018learning}&ResNet-12& 62.45$\pm$\footnotesize{0.98} & 76.11$\pm$\footnotesize{0.69}\\
		AM3~\cite{feat}&ResNet-12& 73.60$\pm$\footnotesize{0.00} & 79.90$\pm$\footnotesize{0.00}\\
		FEAT~\cite{feat}&ResNet-12& 68.87$\pm$\footnotesize{0.22} & 82.90$\pm$\footnotesize{0.15}\\
		Mul. ProtoNets~\cite{mmproto}&ResNet-12& 75.01$\pm$\footnotesize{0.81} & 85.30$\pm$\footnotesize{0.54}\\
		DeepEMD~\cite{ZhangCLS20}&ResNet-12& 75.65$\pm$\footnotesize{0.83} & 88.69$\pm$\footnotesize{0.50}\\	
		AGAM~\cite{agam}&ResNet-12& 79.58$\pm$\footnotesize{0.25} & 87.17$\pm$\footnotesize{0.23}\\	
		Dual TriNet~\cite{DualTriNet}&ResNet-18& 69.61$\pm$\footnotesize{0.46} & 84.10$\pm$\footnotesize{0.35}\\
		Baseline~\cite{chen2019closer}&ResNet-18& 65.51$\pm$\footnotesize{0.87} & 82.85$\pm$\footnotesize{0.55}\\
		Baseline++~\cite{chen2019closer}&ResNet-18& 67.02$\pm$\footnotesize{0.90} & 83.58$\pm$\footnotesize{0.54}\\
		\midrule  
		\textbf{ASL} &ResNet-12& \textbf{82.12$\pm$\footnotesize{0.14}} &\textbf{89.65$\pm$\footnotesize{0.11}} \\
		\bottomrule
		\toprule
		\multirow{2}{*}{\textbf{Model}}&\multirow{2}{*}{\textbf{Backbone}}& \multicolumn{2}{c}{\textbf{SUN}}
		\\		\cline{3-4}
		&& \textbf{1-shot} & \textbf{5-shot}\\
		\midrule
		MatchingNet~\cite{vinyals2016matching}&Conv-64F&  55.72$\pm$\footnotesize{0.40} & 76.59$\pm$\footnotesize{0.21}\\
		ProtoNets~\cite{snell2017prototypical} &Conv-64F&  57.76$\pm$\footnotesize{0.29} & 79.27$\pm$\footnotesize{0.19}\\
		Relation Nets~\cite{sung2018learning}&Conv-64F& 49.58$\pm$\footnotesize{0.35} & 76.21$\pm$\footnotesize{0.19}\\
		Comp.~\cite{comp}  &ResNet-10&45.90$\pm$\footnotesize{0.00} & 67.10$\pm$\footnotesize{0.00}\\
		AM3~\cite{feat}&Conv-64F& 62.79$\pm$\footnotesize{0.32} & 79.69$\pm$\footnotesize{0.23}\\
		AGAM~\cite{agam}&Conv-64F& 65.15$\pm$\footnotesize{0.31} & 80.08$\pm$\footnotesize{0.21}\\
		\midrule  
		\textbf{ASL} &Conv-64F& \textbf{66.17$\pm$\footnotesize{0.17}} &\textbf{80.91$\pm$\footnotesize{0.15}} \\
		\bottomrule
	\end{tabular}
\end{table}
\begin{table}[t]
	\caption{Average accuracy comparison before and after incorporating ASL into existing methods. Backbone: Conv-64F.}
	\centering
	\begin{tabular}{ccc}
		\toprule
		\label{inco}	
		\textbf{Method}  	
		&\textbf{5-way 1-shot} & \textbf{5-way 5-shot}\\
		\midrule	
		Matching Nets~\cite{vinyals2016matching}&  61.16$\pm$\footnotesize{0.89} & 72.86$\pm$\footnotesize{0.70}\\
		\textbf{Matching Nets + ASL}&  \textbf{69.13$\pm$\footnotesize{0.18} (\textit{+7.97})}  & \textbf{74.94$\pm$\footnotesize{0.11} (\textit{+2.08})} \\
		\midrule
		ProtoNets~\cite{snell2017prototypical}&  51.31$\pm$\footnotesize{0.91} & 70.77$\pm$\footnotesize{0.69}\\
		\textbf{ProtoNets + ASL}&  \textbf{74.82$\pm$\footnotesize{0.17} (\textit{+23.51})}  & \textbf{80.15$\pm$\footnotesize{0.12} (\textit{+9.38})} \\
		\midrule
		Relation Nets~\cite{sung2018learning}& 62.45$\pm$\footnotesize{0.98} & 76.11$\pm$\footnotesize{0.69}\\
		\textbf{Relation Nets + ASL}& \textbf{67.02$\pm$\footnotesize{0.19} (\textit{+4.57})} & \textbf{79.73$\pm$\footnotesize{0.13} (\textit{+3.62})}\\
		\bottomrule
	\end{tabular}
\end{table}

\section{EXPERIMENTS}
In this section, the effectiveness of ASL is verified by various experiments. Two standard few-shot classification datasets CUB~\cite{welinder2010caltech} and SUN~\cite{sun} are selected to compare the performance of our approach with previous few-shot recognition methods.

\begin{table*}[t]
	\begin{minipage}{0.35\linewidth}
		\caption{Ablation study on our model. Backbone: ResNet-12.}
		\centering
		\begin{tabular}{ccc}
			\toprule
			Method     
			&5-way 1-shot & 5-way 5-shot\\
			\midrule	
			w/o VAP &73.23$\pm$\footnotesize{0.17}&82.59$\pm$\footnotesize{0.13}\\
			w/o CAM &69.86$\pm$\footnotesize{0.19} &86.64$\pm$\footnotesize{0.13} \\
			w/o PSAM &79.21$\pm$\footnotesize{0.17} &87.87$\pm$\footnotesize{0.13} \\
			w/o VAP \& AVAM &69.60$\pm$\footnotesize{0.19}  &79.83$\pm$\footnotesize{0.23} \\
			Not using attributes & 69.98$\pm$\footnotesize{0.19} & 86.48$\pm$\footnotesize{0.11} \\
			Using all-0 attributes & 69.76$\pm$\footnotesize{0.17} & 84.54$\pm$\footnotesize{0.11} \\
			ASL &\textbf{82.12$\pm$\footnotesize{0.14}} &\textbf{89.65$\pm$\footnotesize{0.11}} \\
			\bottomrule
		\end{tabular}
	\end{minipage}
	\begin{minipage}{0.35\linewidth}
		\caption{Influence of kernel combination. Backbone: ResNet-12.}
		\centering
		\begin{tabular}{ccc}
			\toprule
			\label{}
			Kernel Size     
			&5-way 1-shot & 5-way 5-shot\\
			\midrule	
			3&80.65$\pm$\footnotesize{0.16} &86.61$\pm$\footnotesize{0.12} \\
			5&80.49$\pm$\footnotesize{0.15} &87.05$\pm$\footnotesize{0.12} \\
			7&78.30$\pm$\footnotesize{0.17} &86.69$\pm$\footnotesize{0.11} \\
			9&76.37$\pm$\footnotesize{0.17} &86.03$\pm$\footnotesize{0.12} \\
			(3, 5, 7) &80.78$\pm$\footnotesize{0.17}&87.99$\pm$\footnotesize{0.13}\\
			(5, 7, 9)&80.89$\pm$\footnotesize{0.18} &88.33$\pm$\footnotesize{0.10} \\
			(3, 5, 7, 9)&\textbf{82.12$\pm$\footnotesize{0.14}} &\textbf{89.65$\pm$\footnotesize{0.11}} \\
			\bottomrule
		\end{tabular}
	\end{minipage}
	\begin{minipage}{0.28\linewidth}
		\caption{Influence of weighting factor $\alpha$. Backbone: ResNet-12.}
		\centering
		\begin{tabular}{ccc}
			\toprule
			\label{}
			$\alpha$ &5-way 1-shot & 5-way 5-shot\\
			\midrule
			0& 76.93$\pm$\footnotesize{0.15} &83.60$\pm$\footnotesize{0.11} \\
			0.001&77.85$\pm$\footnotesize{0.14}&85.91$\pm$\footnotesize{0.12} \\
			0.01&78.25$\pm$\footnotesize{0.18}&86.61$\pm$\footnotesize{0.13}  \\
			0.1&80.24$\pm$\footnotesize{0.18} &87.69$\pm$\footnotesize{0.12} \\
			0.5 &80.45$\pm$\footnotesize{0.16} &87.15$\pm$\footnotesize{0.13} \\
			1.0 &\textbf{82.12$\pm$\footnotesize{0.14}} &\textbf{89.65$\pm$\footnotesize{0.11}} \\
			2.0&79.01$\pm$\footnotesize{0.17}  &86.62$\pm$\footnotesize{0.12} \\
			\bottomrule
		\end{tabular}
	\end{minipage}
\end{table*}

\subsection{Experimental Settings}
Following~\cite{snell2017prototypical,sung2018learning}, we use both shallow four layer convolutional Conv-64F~\cite{vinyals2016matching} and ResNet-12~\cite{sun2019meta} as our backbone network. We train our model from scratch by Adam optimizer~\cite{kingma2014adam} with an initial learning rate $1\times10^{-3}$. For ASL, we set weighting factor $\alpha=1.0$ in all experiments. For VAP, we only use the categories of training set for training. We train our model for 60,000 iterations. For both CUB and SUN, the images are resized to $84\times 84$, and no data augmentations are adopted. During the test stage, we report the top-1 mean accuracy over 10,000 tasks. Note that all experiments in Sec. Ex

\begin{figure}[t]
	\centering
	\includegraphics[height=4cm,width=7.6cm]{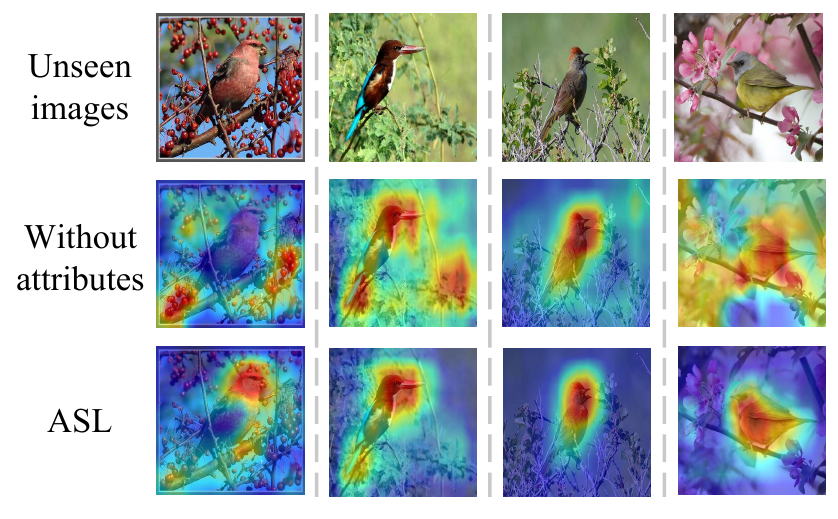}
	\caption{Grad-CAM visualization of four unseen images.}
	\label{cam}
\end{figure}

\subsection{Results and Analysis}
TABLE I lists the recognition results of different methods on CUB and SUN datasets. From the experimental results, we have the following observations:

1) It can be observed that ASL achieves the best performance among all approaches. To be more specific, our model is around 19.2\%/8.1\% better than FEAT~\cite{feat} on CUB with ResNet-12 under 1-shot and 5-shot settings.
	
2) Multi-modal based methods (e.g., Mul. ProtoNets, AM3, Dual TriNet and Comp.) generally outperform classic meta-learning based methods that rely on the single visual modality by a large margin, which validates the effectiveness of using auxiliary semantic information.

3) Among these methods, AM3, Dual TriNet, Comp., and  AGAM only augment the representations of support classes, while query images have no semantic modalities information to enhance representations. ASL generally achieves better performance than these methods, demonstrating that generating and leveraging the attribute information of query images is beneficial to improve representation learning. 

4) The performance improvement of the multi-modal based methods on CUB is greater than that on SUN. This may be because there are more attribute categories in CUB, which can better guide visual attention.

\subsection{Discussion}
\textbf{Pluggability of ASL.} To verify the effectiveness of our proposed ASL, we embed it into three classic metric-based approaches. Table II shows the gains obtained by incorporating ASL into each approach on CUB, and for all three approaches, incorporating ASL leads to a significant improvement. These experimental results verify that our model is a flexible plug-and-play method.

\textbf{Framework Design.} As shown in Table III, each module in ASL has a significant contribution to performance improvement. 1) Removing the VAP drops the performance by 10.8\%/7.9\% on CUB under 1-shot/5-shot settings respectively, which attests that the attribute generation task brings extra useful information and significant improvement. 2) Removing either of the CAM and the PSAM leads to 32.0\%/3.5\% performance drops on the 1-shot setting. This demonstrates that using both attention modules simultaneously can captures more useful information. 3) To prove the effectiveness of our whole model ASL, we also remove both VAP and AVAM. Removing these two modules drops the performance by 15.2\%/11.0\% on 1-shot/5-shot settings respectively. 4) If we only use visual features, the improvement brought by pure visual attention mechanism is limited.

\textbf{Influence of kernel combination.}
We select various kernels and their combinations to conduct experiments on CUB dataset. As shown in Table IV, it is better to use a combination of different kernels. A possible explanation is that different attribute features of objects correspond to different sizes in images, and using a single size cannot precisely locate attribute features. 

\textbf{Influence of weighting factor $\alpha$.} $\alpha$ is a pre-defined parameter that influences the recognition performance. Experimental results on CUB dataset are shown in Table V. It can be seen that when we set $\alpha$ to 1,  ASL can achieve the best performance. 

\textbf{More discussions.} We further inserted AVAM after each ResBlock and achieve 83.42\%/90.32\% accuracy on CUB under 1-shot/5-shot settings with ResNet-12 respectively, indicating that using AVAM at different layers can provide more useful auxiliary information.

\textbf{Visualization.} To make an intuitive understanding of our ASL, Fig. 3 visualizes the Grad-CAM~\cite{gradcam}.
It can be seen that ASL improves the recognition performance by introducing additional attribute vectors to help the model focus on more representative local features.

\section{CONCLUSION}
In this letter, we argue that auxiliary semantic modalities are necessary for query images. We propose an attribute-shaped learning method to generate corresponding attributes through visual features and learn more discriminative visual features. Experimental results illustrate the encouraging performance of our ASL, which has achieved competitive results with other state-of-the-art few-shot learning methods.

\bibliographystyle{IEEEbib}
\bibliography{ASL}

\end{document}